# *In memoriam* Maurice Gross

**Éric Laporte**

Institut Gaspard-Monge, University of Marne-la-Vallée/CNRS
5, bd Descartes, F 77454 Marne-la-Vallée CEDEX 2
eric.laporte@univ-mlv.fr

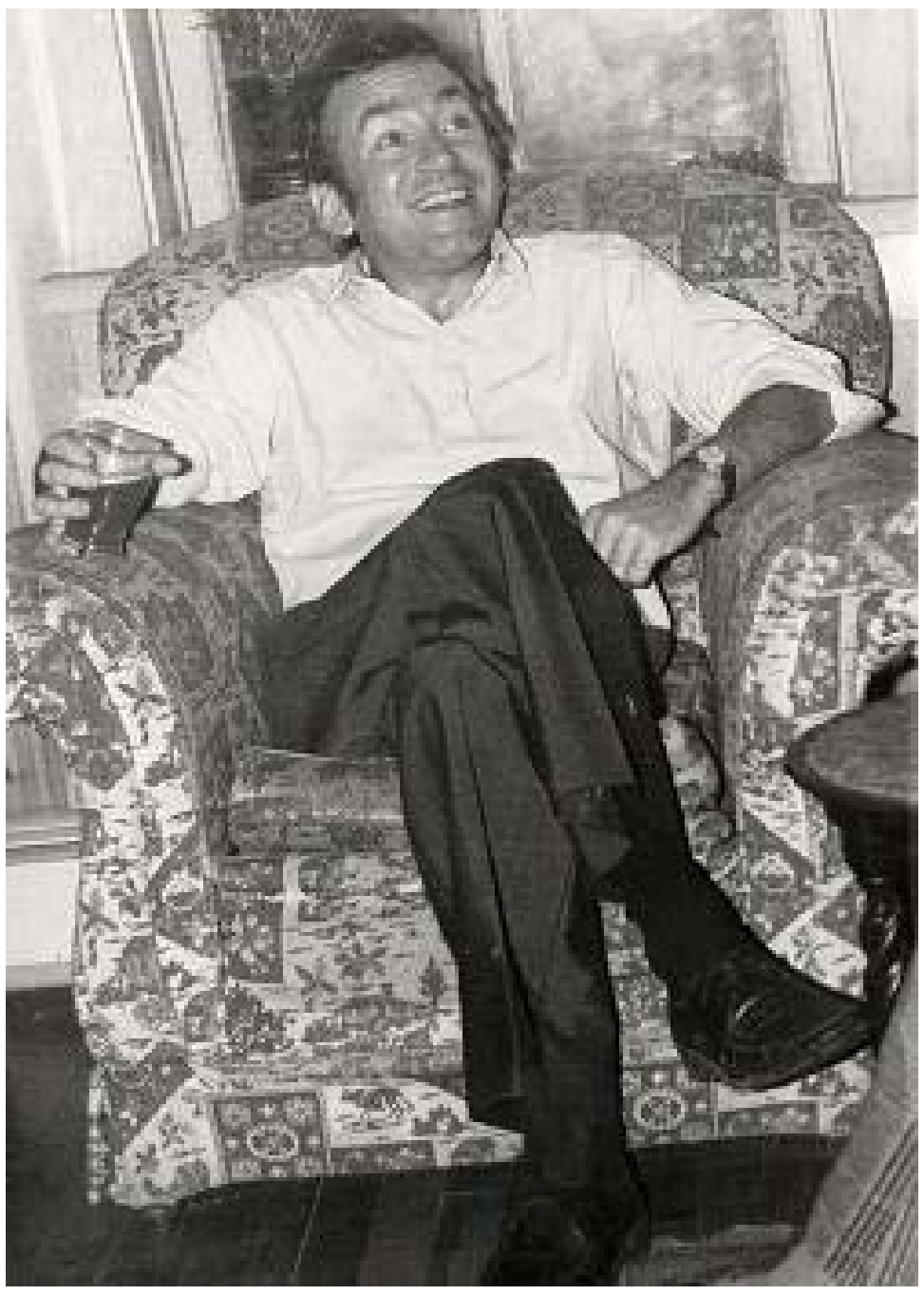

Photo: Jacques Labelle

Maurice Gross (1934-2001) was both a great linguist and a pioneer in natural language processing. This article is written in homage to his memory[1].

Maurice Gross is remembered as a vital man surrounded by a lively group, constituted by his laboratory, the LADL[2], and by his international network, RELEX. He and his group always resolutely steered away from any kind of abstemiousness or abstinence. He selected three of his first collaborators in 1968 from among his bistrot pals of longstanding. Judging by their later scientific production, the selection was done with sharp judgment (e.g. Guillet, Leclère, 1992; Meunier, 1999). A convivial atmosphere, picnics, drinks at the lab and other revelries were the hallmarks of his group — though he has been perceived, on other occasions, as a tyrannical father.

As a linguist, Maurice Gross contributed to the revival of formal linguistics in the 1960s, and he created and implemented an efficient methodology for descriptive lexicology. As specialist of natural language processing (NLP), he was also a pioneer of linguistics-based processing.

## 1. Linguistics

The best-known theory of Maurice Gross is that the description of idiosyncratic properties of lexical elements, i.e. the description of the lexicon, is an essential part of the description of the syntax and semantics of a language. He undertook to implement Zellig Harris' theory of syntax, which in fact also includes a good deal of semantics, within a well-crafted description.[3]

### 1.1. Results: lexicons and grammars

His work comprises not only a vast array of books and articles, among the most significant being those of 1975 and 1981, but also a large data set: tables of syntactico-semantic properties of thousands of lexical entries in French — not a common achievement among linguists. As an illustration of this type of resource, we reproduce an excerpt of a table of English verbs (Fig. 1).

---

[1] I thank Christian Leclère for our discussions and his suggestions, that make up a substantial part of this article.

[2] Laboratoire d'automatique documentaire et linguistique, University of Paris 7 and CNRS.

3 There is a smooth continuum from the thought and work of Zellig Harris to that of Maurice Gross, and it is sometimes difficult to differentiate between their respective contributions.

| N0 =: the fact that S | N0 =: for N00 to V W | N00 = N1 | N00 =/= N1 | N0 =: Nhum | | N0 V N1 | N0 V Particle N1 | N1 =: Nhum | N1 =: Npl + Ncoll | N1 =: Np of Nhum | N1hum V H-self that S | N1hum be V-en that S | N0 be V-ing (to + for) N1 | N0 be V-ble Prep N1 | N0 be V-(ive + ic + icial) Prep N1 | N0 be V-(ous + ful + some) Prep N1 | N0 be V-(ant + ory) Prep N1 |
|---|---|---|---|---|---|---|---|---|---|---|---|---|---|---|---|---|---|
| + | + | - | + | - | blow away | - | + | + | - | - | - | + | - | - | - | - | - |
| + | + | + | + | - | blunt | + | - | - | - | + | - | - | - | - | - | - | - |
| + | + | + | + | - | boogle | + | - | + | - | - | - | + | + | - | - | - | - |
| + | + | - | + | + | bolster up | + | + | + | - | + | - | - | - | - | - | - | - |
| + | + | + | + | + | boost | + | - | - | - | + | - | - | - | - | - | - | - |
| + | + | + | + | + | bore | + | - | + | - | - | - | + | + | - | - | - | - |
| + | + | + | + | + | bother | + | - | + | - | - | - | + | - | - | - | + | - |
| + | + | + | + | + | bowl over | - | + | + | - | - | - | + | - | - | - | - | - |
| + | + | + | + | - | box in | - | + | + | - | - | - | - | - | - | - | - | - |
| + | + | + | + | - | brace up | - | + | + | - | - | - | - | + | - | - | - | - |

Fig. 1. A Lexicon-Grammar table (Salkoff, 2002).

Such tables are known as Lexicon-Grammar tables or as LADL tables. The first tables were constructed around 1970, long before comparable lexicons (Gross, 1975). Some are now available on the web (http://infolingu.univ-mlv.fr/english: follow Linguistic Resources). Intimately associated with the name of Maurice Gross, they are an example of the originality, creativity and fecundity of his scientific thought.

Firstly, they are rich repositories of fact. As compared to other syntactico-semantic lexicons, Lexicon-Grammar tables have three salient features:

- senses of verbs are distinguished and represented in separate lexical entries, for example for *John missed his daughter* and *John missed the target*;
- inside a verb entry, different constructions are represented, for example *John missed his daughter* and *Mary was missed by her father*; in that sense, the tables contain the essential elements of a grammar;
- the number of entries is quite comprehensive, e.g. 15,000 French verbs.

Secondly, these tables are also a means of linguistic investigation. During the construction, they are an abundant source of discovery. Once complete, the tables are readable and constitute a tool for further research by the international community of linguists, including for researchers that do not share the theoretical assumptions of their authors.

And they were designed as such. This is an example of the generosity of Maurice Gross. He also used to pass ideas on to students and colleagues, happy when they were developed in publications, and never caring whether he was acknowledged as their originator.

The work completed under his supervision includes a set of morpho-syntactic lexicons called DELA (Courtois, 1990, 2004). We illustrate this resource with a sentence automaton, which is automatically derived from these lexicons and displays the lexical tags of a sequence of words (Fig. 2).

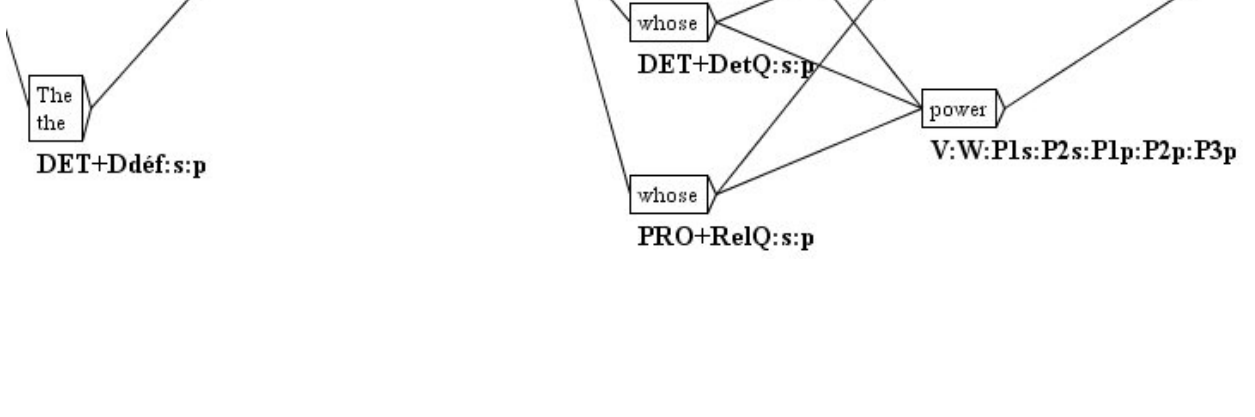

Fig. 2. A sentence automaton with lexical tags in the DELA format.

Thus, the work of Maurice Gross and the RELEX network includes lexicons and grammars encompassing morpho-syntactic and syntactico-semantic levels. Viewed as a whole, it describes both Indo-European languages (French, Italian, Portuguese, Spanish; English, German, Norwegian; Polish, Czech, Russian, Bulgarian; Greek) and others (Arabic; Korean; Malagasy; Chinese; Thai...)

## 1.2. Predicative nouns

The analysis of sentences with a predicative noun and a support verb is a well-known aspect of the theory of Zellig Harris and Maurice Gross (Harris, 1957). Many nouns can function as sentence predicates, playing the same part as verbs, like in

*Human language technologies have an increasing importance*

In such constructions, the noun is accompanied by a support verb, later termed 'light verb' by some writers (e.g. Grimshaw, Mester, 1988). The importance of predicative nouns for natural language processing comes from the fact that most technical nouns are predicative and that in technical texts, many predicates are nouns.

Maurice Gross directed studies on predicative nouns in French and other languages. Large-scale results emerged in the 1970s (Giry-Schneider, 1978) and generally validated the theory. Nonetheless, this theory was felt to be iconoclastic and faced fierce scientific opposition for years. It eventually gained greater acceptance during the 1990s[4]. Resistance to the theory did not influence his opinion because Gross was aware that his opponents' arguments did not stand up to analysis. He knew better than they did which details of the predicative nouns/support verbs model did not work.

Maurice Gross could be obstinate; he liked contradiction, and his critical mind was developed to an uncommon degree. As regards the combination of wine with food — a far weightier issue than predicative nouns — Maurice Gross appreciated some red wines with fish and some white wines with cheese, and he enjoyed demonstrating his irritation and amusement when he received slighted remarks from wine waiters in classier restaurants.

## 1.3. Compounds and idioms

Gross set up a series of studies on compound lexical units (e.g. Freckleton, 1985; Machonis, 1985), breaking with a long tradition that holds that such phrases are exceptions, worth only of anecdotal remarks. Compound lexical units are phrases described as lexical units (Gross, 1986a), as in:

*Cell phones have antennas*

They can be defined by their lack of compositionality and the distributional frozenness of their elements. The findings showed that languages have a large number of compound predicates such as *make ends meet*, and that compound entries are often more numerous than the simple-word entries for the same part of speech. In French, for instance, Maurice Gross indexed 26,000 verbal idioms[5] (Gross, 1982) and 12,000 adverbial idioms (Gross, 1986b).

---

[4] The same can be said for predicative adjectives, as in *Human language technologies are increasingly important*.

He devised the notion of local grammars (Gross, 1997) for semi-frozen phrases and for sequences with frozen behaviour inside a specific domain, like *cloudy with sunny periods*.

The term 'multi-word units' (Glass, Hazen, 1998) is more recent. It groups support-verb constructions, compound lexical units, semi-frozen phrases and collocations.

### 1.4. Methodological legacy

Maurice Gross contributed centrally to the construction of rigorous empirical methods for syntactic and semantic description, borrowing fundamental notions from experimental sciences like biology and physics. This is probably his greatest achievement, but it obviously accrued still further weight from his parallel descriptive work, which showed that his methods were applicable to real, unrestricted data. Historically, this methodological work was part of the international movement in scientific, formalised linguistics that took place in the 1960s. In 1968, Maurice Gross also contributed to the foundation of the Linguistics Department of the University of Vincennes, and was the editor of *Langages* 9, a volume of articles by major contemporary players such as Yehoshua Bar-Hillel, Noam Chomsky, Maurice Gross, Zellig Harris and Marcel-Paul Schützenberger.

Resorting to experimental sciences for linguistic methodology may sound strange to many linguists, even today, but it is justified by the fact that the only possible source of knowledge in descriptive linguistics is empirical observation. The central epistemological notion there is the reproducibility of experiments, i.e. that the results of observation should not depend on the observers. Obviously, the objective of reproducibility is particularly difficult to attain in linguistics. In addition, any steps taken in order to ensure reproducibility should be compatible with the objective of actual large-scale description. Maurice Gross advocated a subjective method with a collective control: empirical observations are performed through introspection by a team of native-speaker linguists. This is how the Lexicon-Grammar tables of French verbs were constructed.

Some argue that sufficient reproducibility can only be ensured by way of total objectivity. However, whatever precautions can be taken in order to ensure the objectivity of results (blind experimentation, exclusive resort to attested language productions…), they are in practice incompatible with the large-scale description of a language.

## 2. Natural language processing

### 2.1. Construction of language resources

Maurice Gross designed and implemented a set of methods and tools for the manual elaboration of language resources of quality.

As compared to common practices in language resource construction and management, the quality of resources constructed with these methods can be characterised by four features:

- Coverage is large, which is obtained by systematic browsing of segments of the language (parts of speech, for instance)
- Representation is formal, e.g. encoded, to permit computational exploitation for natural language processing purposes
- Models for different languages are parallel
- Information is particularly accurate, since it is based on manual analyses of introspective or attested data by trained linguists, and errors are corrected directly when discovered. (In resources automatically derived from rules or corpora, the information provided is approximate; when an error in a resource is discovered, methods to correct it are not sure to be successful and can bring about further errors.)

This level of quality is obtained, as said, through essentially manual construction and maintenance. This implies two additional features:

- Resources are readable to a high degree, as is exemplified by the two figures above (the relevant information is dense; when necessary, as in the second figure, the display is graphical).
- The accumulation of resources is gradual, thus allowing for the progressive, collaborative construction of large resources from independent elements.

In the application of these methods, Maurice Gross kept a watchful eye on the level of formalization. In complete agreement with Zellig Harris's objective of a formally minimal theory[6], he was very keen on minimalism in abstract complexity. He avoided giving new names to already named notions, as computational linguists usually do[7]. He carefully avoided creating new abstract notions or models, or building complex abstract representations, unless they were really useful. For example, the *VP* node in syntactic trees is useless, and so is the distinction between *PP* and *NP* nodes (the presence/absence and lexical value of the preposition can perfectly be specified without two distinct nodes). Such nodes introduce unnecessary complexity into syntactic trees (Fig. 3), whereas the complexity of the phenomena themselves calls for the simplest possible tools (Fig. 4).

---

[5] Idioms with support verbs, such as *to be heavy-handed*, not included.

[6] This difference between Zellig Harris and Noam Chomsky shows even in their respective conceptions of transformations: for Harris, a transformation relates observable sentence forms; for Chomsky, it is a device to transform a deep structure into a surface structure.

[7] For example, morpho-syntax has been called syntax by so many writers that now they have to call syntax 'deep syntax'.

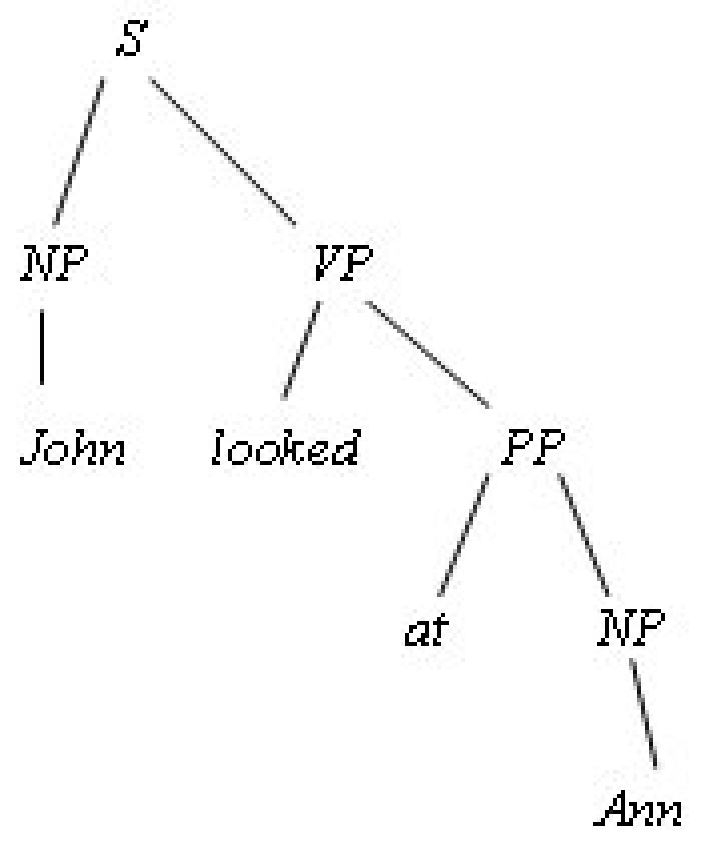

Fig. 3. A syntactic tree with useless nodes.

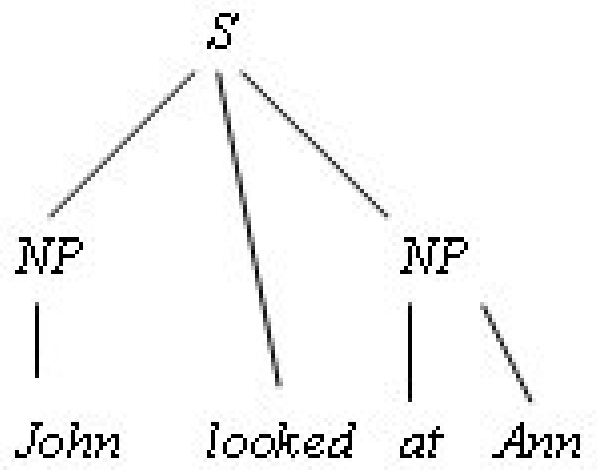

Fig. 4. A simpler syntactic tree.

This position was consistent with his own character. Pragmatic, Maurice Gross liked simplicity and had good sense. He was not easy fooled by apparently attractive and brilliant but poor ideas. He was delighted by the Sokal (1996) hoax against fashionable intellectual discourse.

## 2.2. Linguistics-based language processing

Right from the beginning, Maurice Gross was aware that formal descriptions of natural languages could be applied to language processing. The first opportunity to study linguistics had been offered to him in the 1950s in a machine translation laboratory, during the first period of enthusiasm for machine translation. In the 1960s, he founded and pioneered the concept of linguistics-based NLP, long before the international interest for language resources. In this approach, the construction of NLP programs is based on language resources (lexicons and grammars), which are, in turn, based on descriptive linguistics. He began to apply this concept in the 1960s and 1970s.

However, statistics-based NLP developed and became dominant in the 1980s: the only language resources required for this approach to NLP are corpora. While specialists of linguistics-based NLP were developing their methods, the bulk of the NLP community was discouraged by the difficulty and cost of such work, but aware that language resources were required for the fulfilment of their objectives. This paradox was called the 'bottleneck of NLP'. The international academic community eventually got an interest for language resources, but only in the 2000s.

In spite of the difficulty and cost of linguistics-based construction of language resources, Maurice Gross largely attained his objectives of quality: coverage, formalisation, standardisation and accuracy. The resources developed by him and under his supervision are used by many laboratories and companies, some of them founded by his followers, and their exploitation still offers huge potential for future research and applications.

## 2.3. Finite-state processing

Maurice Gross's approach to the use of the finite automaton model for NLP was pragmatic and constructive. Soon after the creation of this model, the famous article by Chomsky (1956) was interpreted as a ban on its use in linguistics. There were few transgressions (Woods, 1970). Much later, Maurice Gross (1989a) gave examples indicating that the finite automaton would be a convenient, invaluable tool for syntactic description. Groups related to the University of Helsinki and Rank Xerox were of the same opinion, and it came to be one of the most popular models. The RELEX network began to use finite automata with the graphical editor developed by Silberztein (1993). Empirical studies led to the design of a model of lexicalised syntactic grammar in the form of a large network of finite automata, or 'recursive transition network'.

## 2.4. Rules vs. idiosyncrasies

I myself often heard that Maurice Gross was 'pessimistic' because he did not have confidence in rules where language is concerned.

The notions of rule and idiosyncrasy are symmetrical. Rules are generic statements, such as 'nouns in *-s* are plural', whereas idiosyncrasies are particular facts, usually about lexical items, such as '*beans* is plural' or '*means* is singular'. Both forms can express formal knowledge, which raises interesting issues: when a rule is right, what is the use of the corresponding idiosyncrasies? and vice versa; should language resources be based on rules or on idiosyncrasies?

In practice, rules tend to be inaccurate because of the ubiquity of chaotic behaviour and exceptions in language, as our example 'nouns in *-s* are plural' suggests. When we devise rules manually, we have a natural tendency to over-estimate their predictive power; when they are automatically derived from corpora, results are approximate. When idiosyncratic data are available, they are more often accurate, i.e. in conformity with observable reality.

On the other hand, rules can be automatically inferred, whereas accounts of idiosyncratic facts have to be recorded manually, a work which obviously requires time, skill and effort, and is therefore more costly.

In this trade-off between accuracy and cost, software engineers tend to prefer rules, even if slightly inaccurate, because they feel that rules achieve a higher level of automation than collections of particular facts. In addition, they unanimously say that the manual construction of language resources is difficult, and often that it is tedious. However, this is not a specialist's opinion, since they seldom have the opportunity to engage in such tasks. On the contrary, Maurice Gross took evident pleasure in assembling and disassembling mechanisms of language.

Linguists are obviously interested in discovering rules. In good scientific practice, such discovery requires observing facts as completely as possible, but most linguists do not check large numbers of idiosyncratic facts, and they never say explicitly why: do they also consider studies on lexis as tedious? or as a second-rate or shameful work? In any case, such studies have been consistently neglected.

Back to the 1960s, when Maurice Gross undertook his work on lexis, issues about rules and idiosyncrasies were being discussed among linguists. Harris (1951) admitted the necessity of representing lexical restrictions: *'We now have sets of morphemic components (and residues), so set up that as nearly as possible all sequences and combinations of them occur (...) It may not be convenient to represent by means of components such limitations of occurrence among morphemes as do not intersect with other limitations involving the same morphemes, or as do not lead to the division of a class into sub-classes clearly differentiated on that basis (...) This is frequently the case for morphemes classes which are grouped together into a general class on the basis of major similarities, but which have small and unpatterned differences in distribution.'*

So did Chomsky (1962): '*There are in fact exceptions to many rules given above, perhaps all. These will have to be separately listed, unless some more general formulation can be found to account for them as well. (…) But discovery of exceptions to grammatical generalizations is of no consequence in itself, except when it leads to an alternative more comprehensive generalization.'*

And again (Chomsky 1965): *'Much of lexical structure is, in fact, simply a classification induced by the system of phonological and syntactic rules. Postal has suggested, furthermore, that there should be a general lexical analysis of lexical items with respect to each rule* R*, into those which must, those which may, and those which cannot be subject to* R*, and has investigated some of the consequences of this assumption. I mention these possibilities simply to indicate that there remain numerous relatively unexplored ways to deal with the problems that arise when the structure of a lexicon is considered seriously. (...) For the present, one can barely go beyond mere taxonomic arrangement of data. Whether these limitations are intrinsic, or whether a deeper analysis can succeed in unraveling some of these difficulties, remains an open question'*.

But none of them shifted from such theoretical observations to the empirical description of a language, including its lexis.

Maurice Gross was the first to consider such a programme as a priority, to draw the consequences and to take up the challenge. His *Transformational Grammar of French* (1968, 1977, 1986b) is a scientific milestone. As Amr Ibrahim (2002) puts it, it clearly shows the limits of rule-based formal accounts of a language, highlighting insurmountable obstacles to the representation of numerous apparently insignificant phenomena, even limited to the scope of simple sentence.

Zellig Harris's theory was a sound basis for such a research programme and he supported Maurice Gross. Meanwhile, Noam Chomsky was insisting that idiosyncratic facts were not worth investigating; Gross (1973, in French, and the famous 1979 version in English) criticised the practices of generative grammar on these grounds.

Maurice Gross' programme is ambitious, because it requires the meticulousness of an entomologist, but he implemented it in a remarkably practical way, because he had the wisdom to recognise priorities and leave the rest for later generations. For example, restrictions on verb tenses are not represented in the present state of Lexicon-Grammar tables of French:

**Une récompense est promise à Luc par Marie*
« A reward is being promised to Luke by Mary »
*Une récompense a été promise à Luc par Marie*
« A reward was promised to Luke by Mary »

Neither are productive pronominal constructions such as:

*Une telle récompense se promet facilement*
« Such a reward is easily promised »

Maurice Gross' decision to launch his programme required audacity, practicality, good sense, and intellectual honesty. He was not afraid of data. He probably described the situation calmly, in such terms as: *'One has to study all sentence types and all types of verbs. Well, let's see what we find.'* This indicates his ability to go to the heart of a problem with precision and lucidity, and to express his conclusions with a brutality and provocation that gave him some Voltairian sense of humour.

He exercised the same humour in his ferocious comments on scientific adversaries, with no regard for political correctness, and through a collection of favoured put-downs such as *'They are all illiterates.' 'He's lightly mentally handicapped.' 'He's a sick man.' 'Their arrogance is equalled only by their incompetence'* and so on.

## 2.5. Corpora vs. lexicons and grammars

In the currently dominant model of NLP, more or less language-independent programs infer information on input text from rules acquired by frequency-based corpus processing.

In practical terms, the extensive use of this model makes language resource management cheaper. The corpora used by the world's programs make up a largely stable set: operations to enhance or extend them or to create new ones are much less frequent than operations making use of them. In addition, constructing corpora, and even tagged corpora, is a smaller investment than constructing lexicons and grammars. The information in state-of-the-art tagged corpora is less complete than the information in state-of-the-art lexicons and grammars. Coverage of lexical elements is poorer. Structural complexity is kept smaller because of the computational limitations of frequency-based inference programs. The annotation of multi-word units (e.g. *make ends meet*) is poorer. Sense distinctions (e.g. between *John missed his daughter* and *John missed the target*) are not made. Indirectly, this policy limits the number of linguists to be educated by universities and hired by companies.

But the price to be paid for this limitation of costs is a lower quality of product. Maurice Gross worked much more on lexicons and grammars than on corpora. Inference-based or statistics-based programs can compensate for part of the difference of information content, but cannot compete with manually constructed lexicons and grammars, because the automatically inferred information is only approximate.

There are two alternative approaches: linguistics-based processing making use of manually constructed lexicons and grammars, and hybrid processing. In the present situation, they are perceived as factors of quality and progress, as is illustrated by the increasing use of compound-word lexicons by language engineering companies and search engines.

This debate is connected to the controversy between corpus linguistics and introspective linguistics.

Introspective linguists use introspection, a subjective human faculty, as one of the sources of linguistic knowledge. Corpus linguists, in contrast, are usually antipathetic to the use of introspective data. Some of them use the term *armchair linguistics* to suggest that introspective linguistics fails to deal with the reality of language, and that this reality is accessible only through attested linguistic productions. However, negative information, such as the inacceptability of some linguistic forms, is also an aspect of the reality of languages, whereas it is accessible only through introspection, and not through corpus exploration. In addition, there is no example of a large-coverage formalized lexicon or grammar constructed on the basis of corpus linguistics.

Maurice Gross was not opposed to the use of corpora, but his practice evolved with time. He began to use corpora only when a professional tool (Silberztein, 1993) was available to explore them. And this tool was based on the lexicons constructed under his supervision. In fact, no work is more inadequately described by the term *armchair linguistics* than his.

His position on the respective potential of corpora, lexicons and grammars conflicted both with the statistics-based approach to NLP and with the positions of strict corpus linguists. Among the reasons why his scientific arguments could not persuade them, we find a number of extra-scientific reasons, and in particular, sociological explanations. The opposition between statistics-based and linguistics-based NLP is founded on the division between engineers and linguists, a very strong cultural gap; and the opposition between introspective and corpus linguistics had many features of a generation gap. Human factors like these can have such a power on people's (and even scholars') minds that they can prevent them from accepting perfectly rational argument. As an independent thinker, Gross was not impressed by a consensus based partly on non-scientific reasoning.

This particular personal trait showed in the way he educated his students. When he talked with them — and he invested a lot of time in the education of some of them — he focused on scientific matters, leaving them to find themselves training in more mundane skills or political strategies.

## 2.6. A few scientific fashions

Needless to say, Maurice Gross did not make concessions to unjustified scientific fads and fashions. His positions indicate a remarkably long-term view and intuition. Here are three examples of scientific fashions he struggled against, with solid scientific arguments on his side.

### 2.6.1. Machine translation without a lexicon

Not only did Maurice Gross (1972) insist that lexicons are required for machine translation, but also that the lexicon is the first element that should be constructed, because virtually all other components depend on it. Machine translation is now regularly taken as an example of a field that requires lexicons. The modern notion of translation memories is just a variant of the notion of lexicon.

### 2.6.2. Deriving lexicons from conventional dictionaries

The idea of deriving lexical resources from conventional dictionaries (called machine-readable dictionaries, MRDs) became amazingly popular in the end of the 1980s. It appeared as an alternative to the costly task of manually constructing lexical resources. Maurice Gross (1989b) denounced this trend as an illusion and a wrong answer to a real problem, showing that information is much less formal and systematic in dictionaries written for human readers than in the resources required for NLP. This prediction was confirmed years later: the results of these researches were disappointing (Ide, Véronis, 1993).

### 2.6.3. Small tagsets

The most commonly used tagsets are small: they have between 15 to 100 tags, which is what is needed to encode morpho-syntax, lemma excluded, in many languages. The most basic operations on tags are word annotation (assigning tags to words) and ambiguity resolution (removing wrong tags). The latter problem is far more challenging than the former. Both depend heavily on the size of the tagset.

Maurice Gross used to warn that larger tagsets are required for handling lemmata or syntactico-semantic information. However, since the 1990s, most of the field seems to have considered that one should keep on using small tagsets as long as technology to solve lexical ambiguity is not available. This view aims at limiting the complexity of tagged text, which is used as input for further operations. But it also severely limits the investigation of approaches to annotation as well as approaches to ambiguity removal.

International interest for larger tagsets has progressively increased during the last 5 years.

### 2.6.4. Comment

Superficial scientific fashions appear from time to time, and can endure, even though not always a sign of progress. They may lead research in wrong directions, and thus cause waste. This is likely, because research outcomes are not known beforehand. But it is interesting to notice cases where failure could have been avoided by paying attention to scientific arguments presented by an outstanding scholar.

The current mechanisms of scientific democracy, and in particular peer review, regulate research agendas and undoubtedly represent progress, but they cannot prevent undesirable fashions: these checks and balances developed progressively during the career of Maurice Gross, and were quite well installed even 20 years ago, but they failed to prevent short-sighted views from invading the field for years after.

## 3. Institutional legacy

In addition to his works, Maurice Gross's legacy includes an enduring research infrastructure, made up of the following elements:

- an informal network of specialists of linguistics-based language resources, RELEX.
- a series of International Conferences on Lexis and Grammar, held since 1981 (Liverpool 2005, Palermo 2006).
- an international journal with a selection committee, *Lingvisticæ Investigationes*, published by Benjamins.
- a model and a format for lexical resources implemented in numerous languages: the DELA format (Courtois, 1990). The model, the format and the resources were the main source of inspiration for the development of standards in lexical resources by the Genelex project, and indirectly for later standards.

## Conclusion

Through the originality of his legacy and his long-term vision, Maurice Gross is an impressive scientist. Most of his innovations, even those dating back to the 1960s and 1970s, are now becoming increasingly accepted by the scientific community around the world. It is all the more interesting to notice how much and for how long they met with resistance when he first expressed them.

Maurice Gross is worth reading now and for some time to come. His work is more innovative than that presented at many high-profile conferences and journals this year. And it will be so next year.